\documentclass[10pt,twocolumn,letterpaper]{article}

\usepackage[pagenumbers]{iccv} 

\definecolor{iccvblue}{rgb}{0.21,0.49,0.74}
\usepackage[pagebackref,breaklinks,colorlinks,allcolors=iccvblue]{hyperref}
\usepackage{graphicx}
\usepackage{tabularray}
\usepackage{amsmath}
\usepackage{multirow}
\usepackage{wrapfig}
\usepackage[capitalize]{cleveref}
\usepackage{subcaption}
\usepackage{tabularx}
\usepackage{colortbl}
\usepackage{multicol}
\definecolor{darkgreen}{RGB}{0, 154, 85}
\newcolumntype{Y}{>{\centering\arraybackslash}X}

\definecolor{myred}{RGB}{192, 0, 0}
\definecolor{myyellow}{RGB}{255, 192, 0}
\usepackage{soul}
\usepackage{threeparttable}

\def\paperID{6721} 
\def\confName{ICCV}
\def\confYear{2025}

\title{
    \raisebox{-.15\height}{\includegraphics[width=0.033\textwidth]{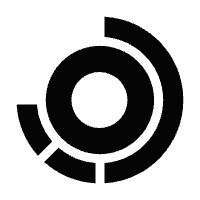}} \textit{TurboTrain}: Towards Efficient and Balanced Multi-Task Learning\\
    for Multi-Agent Perception and Prediction
}

\author{
Zewei Zhou\thanks{Equal contribution. \texttt{\{zeweizhou, sethzhao506\}@ucla.edu}}, \
Seth Z. Zhao\footnotemark[1], \
Tianhui Cai, \
Zhiyu Huang\thanks{Corresponding author. \texttt{zhiyuh@ucla.edu}}, \
Bolei Zhou, \
Jiaqi Ma \\ [0.1cm] 
University of California, Los Angeles\\
\small {\href{https://github.com/ucla-mobility/TurboTrain}{\tt https://github.com/ucla-mobility/TurboTrain}}
}

\begin{document}
\maketitle
\begin{abstract}
End-to-end training of multi-agent systems offers significant advantages in improving multi-task performance. However, training such models remains challenging and requires extensive manual design and monitoring. In this work, we introduce \textbf{TurboTrain}, a novel and efficient training framework for multi-agent perception and prediction. TurboTrain comprises two key components: a multi-agent spatiotemporal pretraining scheme based on masked reconstruction learning and a balanced multi-task learning strategy based on gradient conflict suppression. By streamlining the training process, our framework eliminates the need for manually designing and tuning complex multi-stage training pipelines, substantially reducing training time and improving performance. We evaluate TurboTrain on a real-world cooperative driving dataset, V2XPnP-Seq, and demonstrate that it further improves the performance of state-of-the-art multi-agent perception and prediction models. Our results highlight that pretraining effectively captures spatiotemporal multi-agent features and significantly benefits downstream tasks. Moreover, the proposed balanced multi-task learning strategy enhances detection and prediction.
\end{abstract}    
\section{Introduction}

\begin{figure}[t]
    \centering    
    \includegraphics[width=\columnwidth]{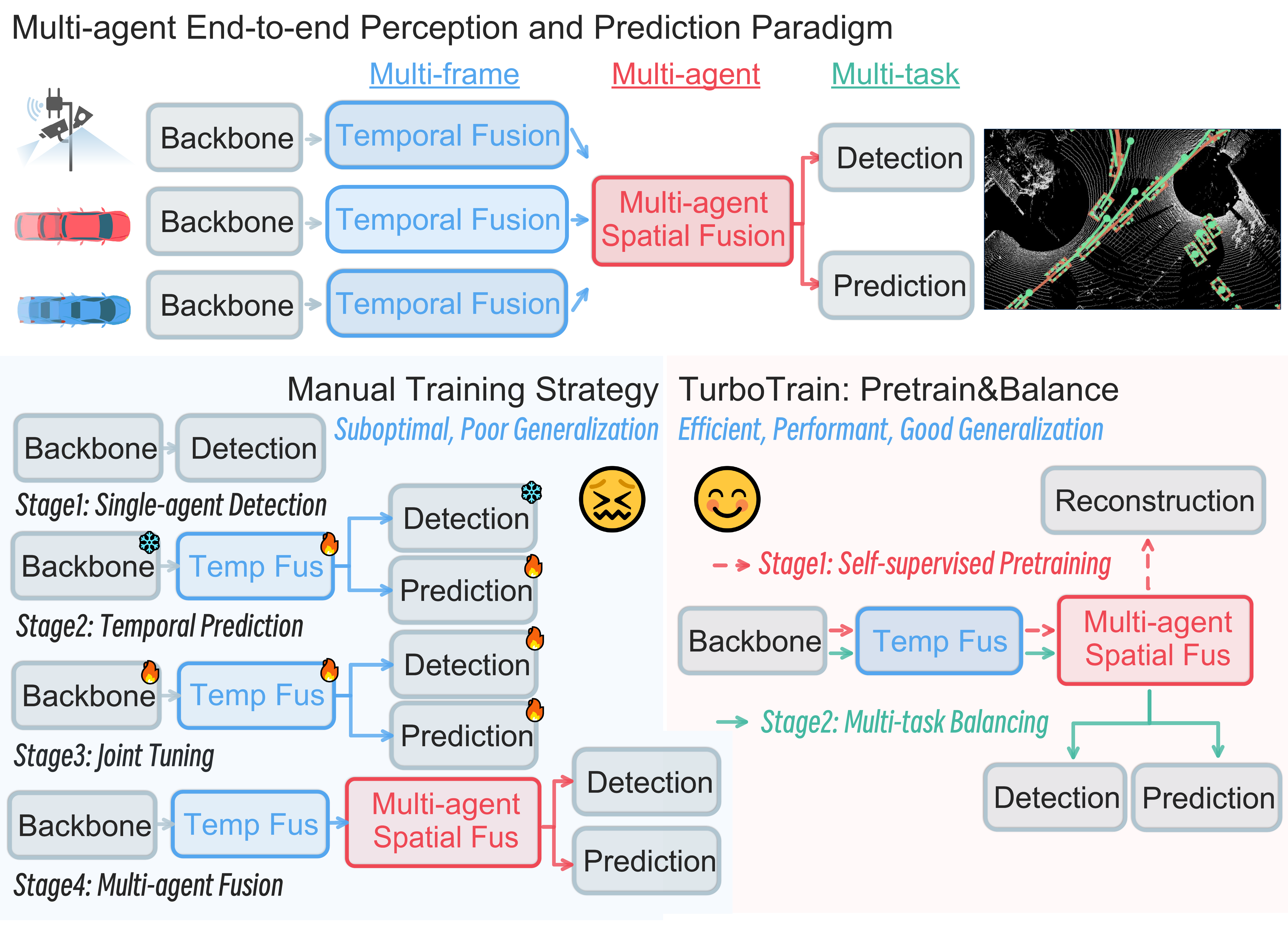}
    \caption{Comparison of manual training strategy and \textit{TurboTrain} paradigm for multi-agent perception and prediction tasks. Unlike the complex manual training strategy, \textit{TurboTrain} is efficient and stable in training with better performance and generalization.}
    \label{fig:idea}
    \vspace{-0.4cm}
\end{figure}

Autonomous driving systems, similar to human drivers, face significant safety challenges in environments with occlusions or limited perception range \cite{liu2023towards, yangjie2024towards, han2024foundation, li2025womdreasoning}. Multi-agent systems mitigate these limitations by leveraging Vehicle-to-Everything (V2X) communication to enable connected and automated vehicles (CAVs) and infrastructure units to share complementary information \cite{xu_opv2v_2022, li_v2x-sim_2022, zhou_comprehensive_2022}. Recent advancements have extended the multi-agent paradigm from single-frame tasks (\eg, detection) \cite{disconet, hu2022where2comm, cooperfuse} to multi-frame temporal tasks (\eg, tracking and prediction) \cite{zhou2024v2xpnp, ruan_learning_2023, wang_v2vnet_2020}. In this context, agents not only aggregate spatial information from neighboring agents but also integrate temporal data across multiple frames, facilitating a comprehensive understanding of the environment.

End-to-end paradigms have gained prominence in single-agent systems by enabling multi-task learning with a unified model and reducing error propagation in decoupled systems \cite{yu2024end, weng2024drive}. However, training end-to-end multi-task frameworks remains challenging and often necessitates multi-stage training to ensure stability \cite{hu_planning-oriented_2023, jiang2023vad, sun2024sparsedrive}. The issues become even more complicated in multi-agent end-to-end systems with multiple tasks, because of two major challenges:
\textit{\textbf{1) Complex Spatiotemporal Features Across Multiple Agents and Frames.}} Efficiently fusing shared features across multiple agents and temporal frames is a significant challenge \cite{xu_v2x-vit_2022, xiang2024v2x, HEAL2024}. Simple spatiotemporal fusion strategies and training paradigms often lead to suboptimal performance or training instability due to complex feature interactions and task conflicts. To address this, state-of-the-art methods such as V2XPnP \cite{zhou2024v2xpnp} and UniV2X \cite{yu2024end} employ multi-stage training pipelines with task-specific supervision, as shown in \cref{fig:idea}. However, these manual training strategies hinder generalization to new tasks \cite{coreICCV}.
\textit{\textbf{2) Limited Annotated V2X Data for Diverse Tasks.}} Unlike single-vehicle datasets, V2X datasets are costly to collect because of extensive sensor arrays and detailed annotations across multiple vehicles and infrastructure units \cite{xiang2024v2x, yu_v2x-seq_2023, zimmer2024tumtrafv2x, huang2024v2x, v2xrealo}. Scaling supervised learning with annotated data to enhance multi-agent model performance, especially in multiple tasks, is prohibitively costly. Consequently, a critical question arises:
\textit{\textbf{How can we efficiently train a multi-agent, multi-frame, and end-to-end framework to optimize multiple tasks using limited data?}}

To address these challenges, we propose \textbf{\textit{TurboTrain}}, the first efficient multi-task learning paradigm for multi-agent end-to-end autonomy frameworks. \textit{TurboTrain} is designed based on pretraining and balancing, which streamlines the training pipeline and achieves superior multi-task performance. In the pretraining stage, we integrate a dedicated temporal and multi-agent fusion module to enable the model to capture long-term temporal dependencies and effectively fuse spatiotemporal features across multiple agents. In contrast, prior works, such as multi-agent single-frame pretraining \cite{coreICCV, CooPre} and single-agent temporal pretraining \cite{wei2024t, xu20244d, agro2024uno}, fail to fully capture the complexity of multi-agent spatiotemporal features. Additionally, we adopt a dual reconstruction learning strategy that captures both point-level and voxel-level features. Unlike BEV-based reconstruction methods \cite{coreICCV}, which result in significant information loss, our approach preserves fine-grained geometric details for detection tasks and models static objects in sparse voxel representations for prediction tasks. This design ensures that the pretraining stage learns task-agnostic multi-agent features, equipping the model with spatiotemporal understanding essential for downstream performance. Notably, our pretraining method, when combined with simple training, achieves performance comparable to complex manually designed multi-stage training strategies, as demonstrated in \cref{fig:comparison}.

For balanced multi-task learning, while shared features across tasks can enhance overall performance compared to single-task learning \cite{weng2024drive, zhang2022beverse}, conflicting feature demands from different tasks can hinder training convergence \cite{lin2023libmtl, zhang2024e2e}. Following prior works \cite{zhou2024v2xpnp, gu_vip3d_2023, agro2024uno, wang_v2vnet_2020}, we focus on representative multi-task learning in multi-agent systems: cooperative detection (\ie, bounding box regression and classification) and trajectory prediction. We propose a conflict-suppressing gradient-alignment multi-task balancer that dynamically resolves gradient conflicts during training.
Moreover, inspired by the role of randomness in escaping local optima \cite{lin2021reasonable, zhang2020random, hu2024revisiting}, our hybrid training strategy combines balanced gradient descent with free training, to avoid the 1.5× GPU overhead typically incurred by per-step balancing. This not only enhances training efficiency and stability but also achieves superior performance. We summarize our key contributions as follows:
\begin{enumerate}
\item We identify the reason for performance collapse in training multi-agent, multi-frame, and multi-task frameworks and introduce \textit{TurboTrain}, the first efficient and balanced multi-task learning paradigm, comprising task-agnostic self-supervised pretraining and multi-task balancing.
\item We propose a multi-agent spatiotemporal pretraining strategy that enhances feature learning across multiple agents and frames, significantly improving training performance for downstream tasks. Additionally, we develop a gradient-alignment balancer to mitigate task conflicts and a hybrid training strategy to accelerate and stabilize gradient-balanced optimization.
\item We conduct extensive experiments on the real-world \textit{V2XPnP-Seq} dataset and demonstrate that \textit{TurboTrain} significantly improves state-of-the-art methods in multi-agent perception and prediction.
\end{enumerate}
\section{Related Work}
\noindent \textbf{Self-supervised Pretraining for Point Clouds.} 
Self-supervised pretraining has demonstrated significant advancements, especially in general feature learning with limited data \cite{xu2022pretram, li2024pretrain, trajmae, He2021MAE, Feichtenhofer2022STMAE}. Early efforts primarily focused on contrastive learning at the point level with high computational overheads \cite{yin2022proposalcontrast}. Although Bird's-Eye-View (BEV) contrastive learning methods have been proposed \cite{bevcontrast2024, lin2024bevmae}, they still struggle with performance and efficiency compared to reconstructing masked points. Moreover, rendering-based methods are particularly difficult to implement for sparse point clouds with high computational costs \cite{yang2024pred, chen2024trend}. For single-agent pretraining, masked reconstruction has proven effective in learning point clouds 3D representations \cite{min2023occupancymae, yang2023gdmae, tian2023geomae}. Recently, multi-agent pretraining has shown superior performance over single-agent methods by benefiting from an expanded perception field and reduced occlusion. CORE \cite{coreICCV} employed a multi-agent BEV feature reconstruction task to support detection, while CooPre \cite{CooPre} introduced the early fusion in Pretrain. However, most existing approaches in reconstructing masked points focus on single-frame reconstruction and ignore the temporal dependencies between frames. T-MAE \cite{wei2024tmae} leveraged the attention module to capture ego temporal information in reconstruction, but it is limited to using only two frames (less than 0.5 seconds). Capturing multi-agent long-term spatiotemporal features remains an open problem.

\noindent \textbf{Multi-task Learning (MTL).} 
Multi-task learning aims to optimize multiple tasks simultaneously by leveraging shared features to enhance performance and efficiency \cite{hu2024revisiting}. One research direction focuses on learning what to share across tasks by incorporating auxiliary learning modules that determine optimal sharing patterns \cite{xin2024vmt}. However, this approach is often difficult to train and lacks flexibility for adaptation to different tasks \cite{chen2023mod}. Another research stream focuses on balancing the loss weights \cite{lin2023libmtl} or gradients \cite{chen2025gradient} among tasks. Loss-weight-based methods adjust the importance of each task through dynamic weighting \cite{lin2021reasonable, chennupati2019multinet++}, while gradient-based approaches modify gradients during each update step to balance tasks \cite{lin2024smooth, agro2023implicit, yu2020gradient}. These two methods are mathematically unified \cite{lin2023libmtl}, with the latter enabling fine-grained manipulation of conflicts. In autonomous driving, existing MTL methods predominantly concentrate on balancing detection and segmentation tasks within single-frame contexts \cite{yang2024unipad, zhang2024e2e, liang2022effective}. However, challenges associated with balancing multiple temporal tasks, especially in multi-agent systems, remain largely unexplored.

\noindent \textbf{Cooperative Perception and Prediction.}
Extensive studies have focused on cooperative perception using single-frame data \cite{gao2025stamp, disconet,song2024collaborative, xia2024one, AgentAlign}. To effectively utilize temporal information, SCOPE \cite{yang2023spatio} aggregates historical data for enhanced detection, while FFNet \cite{ffnet} and CoBevFlow \cite{wei2023asynchronyrobust} align delayed V2X information by leveraging short-term history. In terms of cooperative prediction, recent studies have shown promising results \cite{yu_v2x-seq_2023, wu2024cmp, ruan_learning_2023, zhang2025co,lei2025cooperrisk}. However, most approaches are reliant on trajectory information and are typically designed as standalone modules after perception. Consequently, there has been growing interest in end-to-end systems that integrate cooperative perception and prediction. V2VNet \cite{wang_v2vnet_2020} utilizes graph neural networks to jointly optimize these tasks. UniV2X \cite{yu2024end} explores an end-to-end pipeline with a straightforward fusion strategy. V2XPnP \cite{zhou2024v2xpnp} introduces a comprehensive spatiotemporal fusion framework for V2X-based perception and prediction. Nonetheless, efficient training strategies for complex multi-agent end-to-end frameworks remain underexplored.

\section{Problem Formulation}
\label{sec:Prelim}
In this section, we first formulate multi-task learning with multi-agent end-to-end frameworks for cooperative perception and prediction. We then highlight the key learning challenges by comparing different training strategies.

\noindent \textbf{Cooperative Multi-task Learning.} 
The cooperative perception and prediction task is formulated as follows. Given a map and historical raw perception sequences ${\mathbf{P}_i^t}, i\in\{1,\cdots,N\}$, $t\in\{1,\cdots,T\}$, from all $N$ agents within the ego agent's communication range, the goal is to detect objects in the current frame and forecast their future trajectories by leveraging map context. The multi-task learning framework aims to simultaneously optimize multiple tasks and mitigate feature conflict, which is formulated as:

\begin{equation}
\min _{\boldsymbol{\theta}^{sh}, \boldsymbol{\theta}^{t}_{d}, \boldsymbol{\theta}^{ta}_{p}} \left(\mathcal{L}_{1}(\boldsymbol{\theta}^{sh},\boldsymbol{\theta}^{ta}_{d}), \mathcal{L}_{2}(\boldsymbol{\theta}^{sh},\boldsymbol{\theta}^{ta}_{p})\right),
\end{equation}
where $\boldsymbol{\theta}^{sh}$ denotes the shared parameters of the multi-task networks, and $\boldsymbol{\theta}^{ta}_{d}, \boldsymbol{\theta}^{ta}_{p}$ represent task-specific parameters for bounding box regression, object classification, and trajectory prediction, respectively. $\mathcal{L}_{i}(\cdot)$ represents the learning objective or loss function for task $i$.

\noindent \textbf{End-to-End Multi-Agent Perception and Prediction.} 

As illustrated in \cref{fig:pipeline}, in a V2X scenario, the ego agent $A_{ego}$ communicates with $N$ cooperative agents $A_i$ within its communication range. Each $A_i$ initially shares metadata, including poses, timestamps, and agent types. All point-cloud data are transformed into the ego agent's coordinate system. We adopt the intermediate fusion strategy proposed in \cite{zhou2024v2xpnp}, which is particularly suitable and communication-efficient for the multi-agent end-to-end paradigm. In this framework, each agent first extracts the feature with a detection backbone in each frame and captures the temporal feature with a temporal module. After compression, single-agent features are shared, and the ego agent performs multi-agent fusion. Then, map features are integrated, and the fused spatiotemporal features are fed into two heads: one for object detection (bounding box regression and classification) and another for trajectory prediction. Further details of our multi-agent end-to-end framework are provided in the Supplementary Material.

\begin{figure}[t]
    \centering    
    \includegraphics[width=\columnwidth]{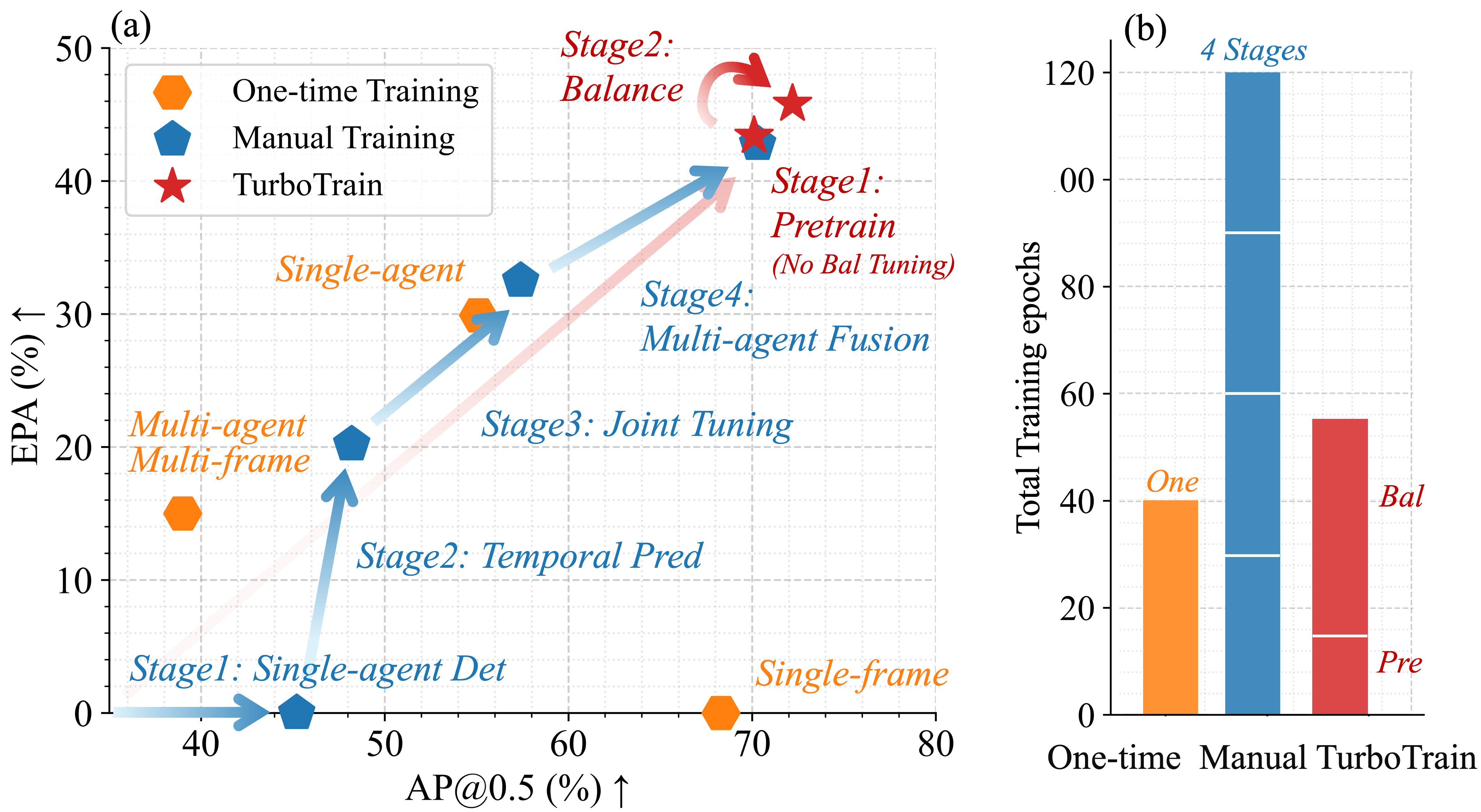}
    \caption{Comparison of one-time training, manual training, and \textit{TurboTrain} on the V2XPnP-Seq-VC dataset using the V2XPnP model. AP evaluates detection performance, while EPA measures joint perception and prediction performance. Dataset and metric details are provided in \cref{sec:exp}. Our \textit{TurboTrain} paradigm demonstrates superior performance and efficiency by significantly reducing training stages and epochs.}
    \label{fig:comparison}
    \vspace{-0.4cm}
\end{figure}

\begin{figure*}[t]
    \centering
    \includegraphics[width=\textwidth]{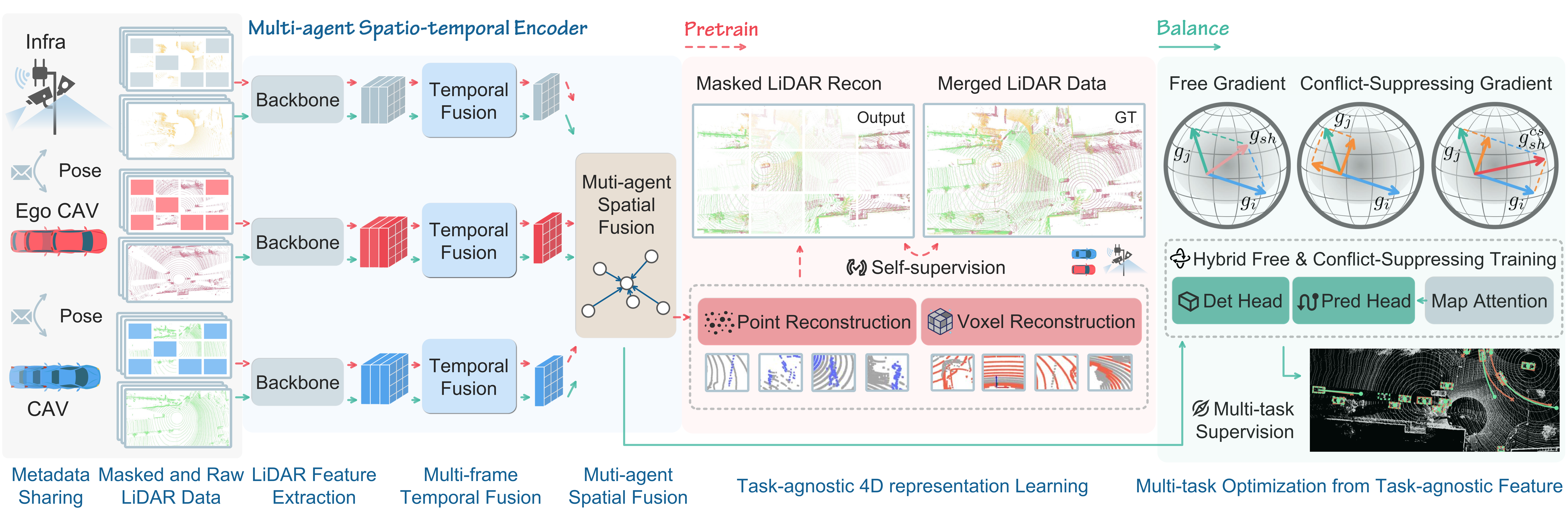}    
    \caption{Illustration of proposed \textit{TurboTrain} paradigm. The \textit{Pretrain} stage focuses on learning a task-agnostic 4D representation within the multi-agent spatiotemporal encoder, which reconstructs the merged masked LiDAR data from multiple agents with masked single-agent data at point-level and voxel-level. The \textit{Balance} stage leverages the pretrained model for multiple downstream tasks and adopts a hybrid training strategy that alternates free and conflict-suppressing gradient steps to efficiently optimize overall multi-task performance.}
    \label{fig:pipeline}
    \vspace{-0.4cm}
\end{figure*}

\noindent \textbf{Training Challenges.} 
We investigate two primary questions: \textit{1) Which component is the most challenging to train and may lead to learning failure?} and \textit{2) How do manual training strategies compare in terms of efficiency, performance, and generalization?} To answer these questions, we compare different training strategies: \textit{1) One-time training} applied to perception and prediction tasks in single-agent and multi-agent settings, including single-frame perception; \textit{2) Manual (multi-stage) training} employed in the SOTA cooperative perception and prediction framework \cite{zhou2024v2xpnp}, which integrates several task-specific optimization phases; and \textit{3) TurboTrain} with pretraining and balancing stages. Our empirical findings reveal that one-time training is viable for single-agent and single-frame settings but fails in multi-agent multi-frame multi-task learning due to its inability to capture complex spatiotemporal dependencies. The manual training strategy mitigates this issue by progressively extending the learning process, starting from single-task detection, then incorporating temporal multi-task learning, and finally scaling to multi-agent multi-frame multi-task learning. While this staged approach enhances stability, it requires numerous training phases and additional epochs (see \cref{fig:comparison}(b)), making it sensitive to task configurations and model architectures, thereby limiting its generalization. To overcome these limitations, our \textit{TurboTrain} framework ensures robust multi-agent multi-task learning by first constructing a task-agnostic 4D spatiotemporal representation in $\boldsymbol{\theta}^{sh}$, followed by balanced optimization of all task-specific parameters $\boldsymbol{\theta}^{ta}_{d}$ and $\boldsymbol{\theta}^{ta}_{p}$. As demonstrated in \cref{fig:comparison}, \textit{TurboTrain} outperforms other training approaches with fewer training epochs, effectively addressing the challenges of multi-agent multi-frame multi-task learning. More illustrations are provided in the Supplementary.
\section{Methodology}
As illustrated in \cref{fig:pipeline}, our proposed \textit{TurboTrain} framework consists of two main components: \textit{1) Pretrain}: multi-agent spatiotemporal pre-training stage designed to learn a task-agnostic 4D representation for the multi-agent spatiotemporal encoder; \textit{2) Balance}: balanced multi-task learning stage aimed at optimizing both perception and prediction from the task-agnostic feature.

\subsection{Multi-Agent Spatiotemporal Pretraining}
\noindent \textbf{Overview.} To effectively learn spatiotemporal features across multiple agents and frames, our \textit{Pretrain} pipeline features two key components: \textit{1) Temporal and multi-agent fusion.} Unlike CooPre \cite{CooPre}, which trains only the 3D backbone using an early fusion strategy, our pipeline trains the 3D backbone, temporal module, and multi-agent fusion module. Given the complexity of spatial-temporal features in multi-agent fusion, this pretraining step is crucial for initializing an effective multi-agent fusion module (see results in \cref{sec:exp}). \textit{2) Multi-agent task-agnostic reconstruction tasks.} In contrast to CORE \cite{coreICCV}, which reconstructs BEV features from multiple agents and may lead to significant information loss due to feature compression, our approach focuses on point cloud and voxel occupancy reconstruction. For point cloud reconstruction, our objective is to restore geometric details of raw points, enabling precise modeling of intricate structures \cite{yan2020pointasnl, hu2019randla}. For occupancy reconstruction, we perform sparse voxel reconstruction to facilitate the model capturing 3D representations of static objects and sparsely populated regions in large-scale environments \cite{hong20224ddsnet, Hong_2021_CVPR}. Besides, we formulate the reconstruction objective to enable the ego agent to reconstruct both its own and collaborating agents' point clouds across spatial spaces and temporal sequences. This design enhances the model’s ability to learn a robust and task-agnostic 4D representation.

\noindent \textbf{Multi-agent Point Cloud Data Collection.}
We define the collection of the point cloud data from the ego agent as $\mathbf{P}_{ego}$ and the cooperative agents as $\mathbf{P}_{coop} = \bigcup_N^{T} \mathbf{P}_i^{t}$. All point cloud data from cooperative agents is projected into the ego agent's coordinate system. This collection of multi-agent point cloud data will serve as a self-supervision signal for the later multi-agent reconstruction objective.

\noindent \textbf{Masked Training Strategy.} 
For each ego agent $A_{ego}$, the raw LiDAR point cloud input is processed through a 3D encoder, yielding BEV features of shape $X \times Y \times C$. A corresponding BEV plane of size $X \times Y$ is used to segment the space into grids $g_{i, j}$. Each LiDAR point $p_k \in \mathbf{P}_{ego} \bigcup \mathbf{P}_{coop}$ is then projected onto a BEV grid based on its $(x, y)$ coordinates. With the inclusion of collaborative agents' point clouds, empty BEV grids are substantially reduced. We randomly apply a high masking ratio to non-empty BEV grids. The unmasked tokens are processed through the 3D encoder, followed by the temporal module to generate the final spatial-temporal features. Note that to ensure training efficiency, the temporal module processes voxel features instead of BEV features. Unlike \cite{wei2024tmae}, which trains a Siamese network using complete historical frames, our method directly trains the 3D encoder and temporal module with fewer point cloud tokens.

\noindent \textbf{Reconstruction Decoder.} 
The decoder is independent of the 3D encoder, allowing flexibility in design. We follow prior works \cite{lin2024bevmae, CooPre, min2023occupancymae} and employ separate lightweight decoders for multi-agent point cloud and occupancy reconstruction. The point cloud reconstruction decoder consists of one convolution layer with masked point clouds as the input and outputs a fixed number of raw points. The occupancy reconstruction decoder consists of three convolutional layers with masked voxels as input and outputs a binary prediction of whether the voxel is occupied.

\begin{figure}[t]
\centering\includegraphics[width=\linewidth]{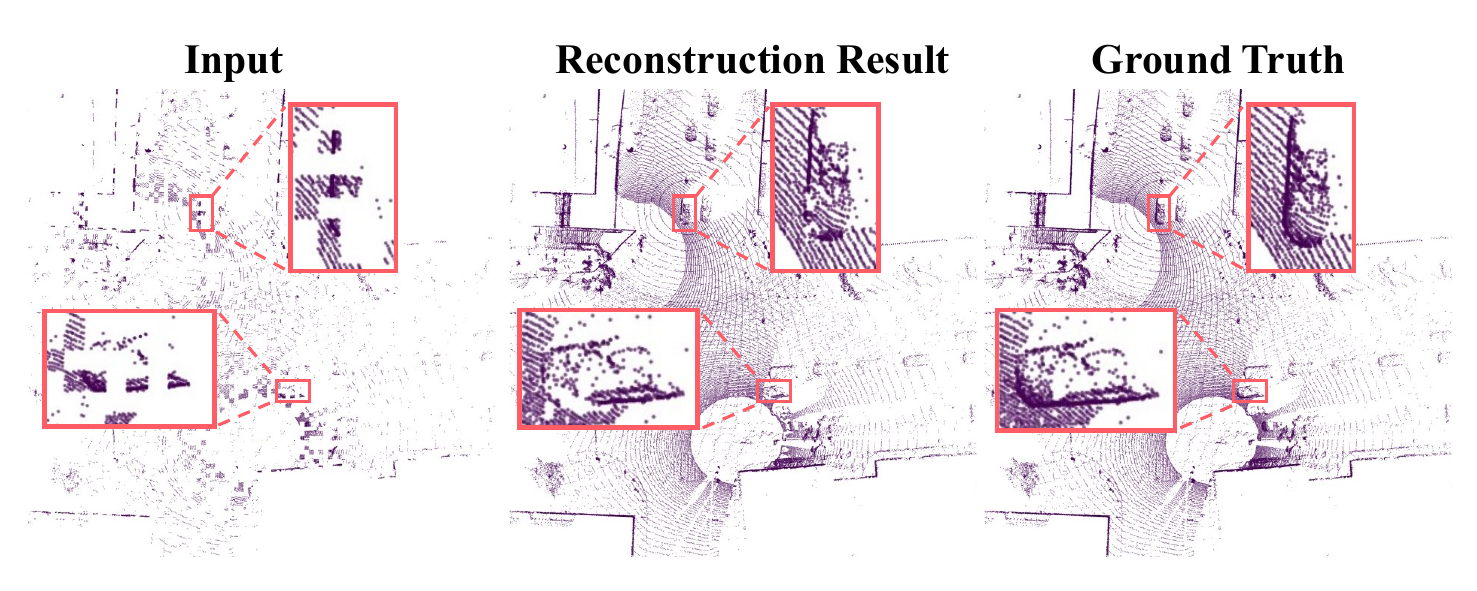}
    \vspace{-0.6cm}
    \caption{Multi-agent point cloud reconstruction visualization of our pretrained V2XPnP model. The model effectively extracts unique LiDAR features from heterogeneous agents, demonstrating a robust understanding of the surrounding 3D environment.}
    \label{fig:recon_vis}
    \vspace{-0.4cm}
\end{figure}

\noindent \textbf{Multi-agent Reconstruction Objective.} 
We leverage highly masked historical and current frame information to reconstruct the current frame. An example of multi-agent reconstruction visualization is shown in \cref{fig:recon_vis}. To learn a task-agnostic LiDAR feature representation, the decoder will perform reconstruction at both point cloud and voxel levels. Specifically, we use a lightweight decoder to generate a fixed number of point clouds for reconstruction, using the Chamfer distance loss as the objective function. For a masked BEV grid $g_{i,j}$, the reconstruction loss between predicted point clouds $\hat{P}$ and ground-truth point clouds $P$ is defined as:
\begin{equation}
\resizebox{.89\hsize}{!}{$\mathcal{L}_{rec}(\hat{P},P)=\frac{1}{|\hat{P}|}\sum\limits_{\hat{p_i}\in\hat{P}}\min\limits_{p_j\in P}\left \| p_i-\hat{p_j} \right \|_2^2+\frac{1}{|P|}\sum\limits_{p_i\in P}\min\limits_{\hat{p_j}\in\hat{P}}\left \| p_i-\hat{p_j} \right \|_2^2$}.
\end{equation}

\noindent Note that the ground-truth point clouds in each masked grid contain a mixture of points from $P_{ego}$ and $P_{coop}$. For the occupancy reconstruction loss $\mathcal{L}_{occ}$, due to the high imbalance of empty voxels, we utilize focal loss for the binary classification problem \cite{lin_focal_2017}.

\subsection{Multi-Task Balancing and Optimization} 
Multi-task learning is a fundamental advantage of end-to-end models. Under the \textit{TurboTrain} paradigm, the \textit{Balance} stage adapts pretrained task-agnostic features to specialized multi-task applications. In this section, we analyze gradient conflicts in multi-task learning and introduce an efficient conflict-suppressing gradient descent strategy to achieve balanced learning across all tasks.

\noindent \textbf{Gradients Conflict.}
Gradient conflicts commonly arise in multi-task learning and have been identified as a primary cause of suboptimal performance and task collapse \cite{chen2025gradient, fernando2023mitigating}. Although multiple tasks share feature representations and parameters within an end-to-end model, their distinct objectives $\mathcal{L}_{i}(\cdot)$ lead to gradient directions $g_i=\nabla_{\theta^{ta}_{i}}\mathcal{L}_{i}(\theta^{ta}_{i})$ that diverge in the solution space. A gradient that benefits one task may inadvertently degrade another, as illustrated in \cref{fig:pipeline}. For instance, the detection task primarily relies on the current frame (with limited contributions from historical frames), whereas the prediction task depends on the entire temporal sequence and cannot be effectively addressed using only the current frame. This divergence in objectives and information dependency brings gradient conflicts. 

\noindent \textbf{Conflict-Suppressing Gradient.} 
Two gradients are considered conflicting if their inner product is negative, indicating that one gradient introduces an opposing component relative to the other. Based on this principle, we detect conflicts by computing the inner product and subsequently removing the conflicting components $\frac{g_{i} \cdot g_{j}}{\left\|g_{j}\right\|^{2}} g_{j}$ (components of $g_{i}$ on the $g_{j}$ direction) when aggregating gradients for the shared parameters $\boldsymbol{\theta}^{sh}$. This ensures that updates are aligned to minimize inter-task interference, thereby enhancing overall model stability and performance.

\noindent \textbf{Hybrid Free \& Conflict-suppressing Training.} 
Computing per-task gradients typically requires multiple backward passes, leading to significant computational overhead and excessive GPU memory consumption, particularly in complex multi-agent end-to-end models. To mitigate this, we integrate free training into our balanced training framework, inspired by the role of randomness in escaping local optima \cite{lin2021reasonable, hu2024revisiting}. By alternating between conflict-suppressing updates and free training, we maintain training stability and optimize model performance without incurring the high computational cost of per-step gradient corrections. Thus, our conflict-suppressing gradient update is described as:
\begin{equation}
\left\{
\begin{array}{ll}
g_{sh}^{cs} = g_{i} + g_{j}, & \text{if } \delta > 0 \text{ or } c_t \geq n, \\
g_{sh}^{cs} = g_{i} + \left(1 - \frac{g_{i} \cdot g_{j}}{\left\|g_{j}\right\|^2}\right) g_{j}, & \text{if } \delta < 0 \text{ or } c_t < n,
\end{array}
\right.
\end{equation}
\begin{equation*}
\delta=g_{i} \cdot g_{j}, \quad c_t = s_t \bmod (n+m),
\end{equation*}
where $g_{sh}^{cs}, \ g_{i}, \ g_{j}$ represent the gradient of aggregated conflict-suppressing for share parameters and task $i,j$, and $\delta$ is the inner product of the gradients. $n, m$ are the gradient steps of free training and conflict-suppressing, and $s_t$ is the current step. Moreover, unlike existing open-source libraries in multi-task learning that lack support for multi-GPU and half-precision training \cite{lin2023libmtl, liu2019end}, our implementation accommodates both, ensuring scalability and efficiency for multi-task learning in multi-agent end-to-end models.
\begin{table*}[t]
\centering
\caption{Cooperative perception and prediction models with \textit{TurboTrain} in VC and V2V collaboration modes on V2XPnP-Seq Dataset. The format of xx$\vert$xx indicates the performance of the models with \textit{TurboTrain} and with the manual training strategy.}
\label{tab:main_results}
\footnotesize
\renewcommand{\arraystretch}{1.1}
\setlength{\tabcolsep}{9pt}
\begin{tabular}{l|l|c|ccc|c}
\toprule[1.1pt]
Dataset & Method & \textbf{AP@0.5} (\%) $\uparrow$ & ADE (m) $\downarrow$ & FDE (m) $\downarrow$ & MR (\%) $\downarrow$ & \textbf{EPA} (\%) $\uparrow$ \\ 
\midrule
\multirow{4}{*}{\shortstack{V2XPnP-Seq-VC\\(\textit{with 2V+2I at most})}} 
  & No Multi-Agent Fusion   & 59.1 {\tiny{\textcolor{darkgreen}{+3.7}}} $\vert$ 55.4  & 1.64 $\vert$ 1.62 &  3.00 $\vert$ 2.93 & 37.7 $\vert$ 35.2 & 35.2 {\tiny{\textcolor{darkgreen}{+1.9}}} $\vert$ 33.3 \\
  & F-Cooper* \cite{fcooper} & 66.3 {\tiny{\textcolor{gray}{-0.2}}} $\vert$ 66.5  & 1.41 $\vert$ 1.32 & 2.56 $\vert$ 2.44 & 34.3 $\vert$ 37.0 & 41.0 {\tiny{\textcolor{darkgreen}{+2.9}}} $\vert$ 38.1 \\
  & DiscoNet* \cite{ffnet}      & 67.2 {\tiny{\textcolor{darkgreen}{+1.2}}} $\vert$ 66.0  & 1.38 $\vert$ 1.39 & 2.59 $\vert$ 2.63 & 36.2 $\vert$ 36.0 & 41.4 {\tiny{\textcolor{darkgreen}{+1.4}}} $\vert$ 40.0  \\
  & CoBEVFlow* \cite{wei2023asynchronyrobust}      & 63.7 {\tiny{\textcolor{darkgreen}{+1.5}}} $\vert$ 62.2  & 1.36 $\vert$ 1.42 & 2.49 $\vert$ 2.59 & 33.0 $\vert$ 32.2 & 41.9 {\tiny{\textcolor{darkgreen}{+1.7}}} $\vert$ 40.2 \\
  & FFNet* \cite{ffnet}      & 70.2 {\tiny{\textcolor{darkgreen}{+2.8}}} $\vert$ 67.4  & 1.43 $\vert$ 1.43 & 2.67 $\vert$ 2.66 & 36.9 $\vert$ 38.0 & 42.1 {\tiny{\textcolor{darkgreen}{+2.3}}} $\vert$ 39.8 \\
  & V2XPnP* \cite{zhou2024v2xpnp}  & \textbf{72.2 \tiny{\textcolor{darkgreen}{+1.4}}} $\vert$ 70.8 & 1.49 $\vert$ 1.57 & 2.75 $\vert$ 3.07 & 35.0 $\vert$ 39.2 & \textbf{45.5 \tiny{\textcolor{darkgreen}{+3.7}}} $\vert$ 41.8 \\
\midrule
\multirow{4}{*}{\shortstack{V2XPnP-Seq-V2V\\(\textit{with 2V})}} 
  & No Multi-Agent Fusion   & 53.5 {\tiny{\textcolor{darkgreen}{+3.1}}} $\vert$ 50.4   & 1.57 $\vert$ 1.79  & 2.94 $\vert$ 3.29  & 38.8 $\vert$ 38.9 & 29.1 {\tiny{\textcolor{darkgreen}{+2.9}}} $\vert$ 26.2 \\
  & F-Cooper* \cite{fcooper} & 63.1 {\tiny{\textcolor{darkgreen}{+4.0}}} $\vert$ 59.1  & 1.59 $\vert$ 1.68 & 2.93 $\vert$ 3.16 & 39.1 $\vert$ 41.2 & 35.1 {\tiny{\textcolor{darkgreen}{+1.9}}} $\vert$ 33.2 \\
  & DiscoNet* \cite{ffnet}      & 66.1 {\tiny{\textcolor{darkgreen}{+3.9}}}  $\vert$ 62.2 & 1.52 $\vert$ 1.56 & 2.84 $\vert$ 2.99 & 39.1 $\vert$ 38.3 & 37.8 {\tiny{\textcolor{darkgreen}{+2.9}}} $\vert$ 34.9 \\
  & CoBEVFlow* \cite{ffnet}      & 65.5 {\tiny{\textcolor{darkgreen}{+2.7}}} $\vert$ 62.8  & 1.69 $\vert$ 1.61 & 3.10 $\vert$ 3.03 & 40.6 $\vert$ 39.8  & 36.8 {\tiny{\textcolor{darkgreen}{+1.2}}} $\vert$ 35.6 \\
  & FFNet* \cite{ffnet}      & 64.2 {\tiny{\textcolor{darkgreen}{+1.9}}} $\vert$ 62.3  & 1.70 $\vert$ 1.68 & 3.11 $\vert$ 3.15 & 39.6 $\vert$ 41.3& 36.4 {\tiny{\textcolor{darkgreen}{+2.8}}}  $\vert$ 33.6  \\
  & V2XPnP* \cite{zhou2024v2xpnp}  & \textbf{68.6} \textbf{{\tiny{\textcolor{darkgreen}{+2.6}}}} $\vert$ 66.3 & 1.85 $\vert$ 1.97 & 3.32 $\vert$ 3.59 & 40.7 $\vert$ 41.4 & \textbf{38.5} \textbf{{\tiny{\textcolor{darkgreen}{+2.3}}}} $\vert$ 36.2\\
\bottomrule[1.1pt]
\end{tabular}
\vspace{-0.2cm}
\end{table*}

\section{Experiments}
\label{sec:exp}
\subsection{Experimental Setup}

\textbf{Evaluation Metrics}.
Following the evaluation protocol of end-to-end perception and prediction in \cite{zhou2024v2xpnp, gu_vip3d_2023}, we employ the Average Precision (AP) for detection task and commonly used prediction metrics, \ie, Average Displacement Error (ADE), Final Displacement Error (FDE), and Miss Rate (MR) within a 2-meter threshold \cite{huang_survey_2022}. Moreover, the prediction accuracy depends on the detection performance, because the false positives and missed objects always mislead the prediction. To evaluate the comprehensive performance of perception and prediction, \textit{End-to-end Perception and Prediction Accuracy (EPA)} metric is adopted and the Intersection over Union (IoU) threshold is set as 0.5.
\begin{equation}
\text{EPA} = \frac{|\hat{S}_{\text{match, hit}}| - \alpha N_{\text{FP}}}{N_{\text{GT}}},
\end{equation}
where $|\hat{S}_{\text{match, hit}}|$ denotes the true positive object number with $FDE <\tau_{EPA}$, while $N_{FP}$ and $N_{GT}$ represent the false positive object number and ground truth object number, respectively. Here, $\alpha$ is a penalty coefficient. A higher EPA value indicates enhanced detection and prediction performance. Following \cite{zhou2024v2xpnp, gu_vip3d_2023}, we set $\tau_{EPA}=2$m, $\alpha=0.5$.

\noindent \textbf{Dataset}.
We evaluate our approach in V2XPnP Sequential Dataset (V2X-Seq) \cite{zhou2024v2xpnp}, the first sequential dataset designed to support various V2X collaboration modes with two CAVs and two infrastructure units. Specifically, we focus on two representative collaboration modes: \textit{1) Vehicle-Centric (VC)}: A CAV agent is the ego agent and communicates with all other CAVs and infrastructure units. \textit{2) Vehicle-to-Vehicle (V2V)}: Communication is restricted to CAVs. In the VC setting, the number of participating agents varies dynamically between 2 and 4, whereas in the V2V setting, it is fixed at two. The dynamic agent number in VC allows us to assess model generalization across different collaboration scenarios. The evaluation protocol remains consistent across all models.

\subsection{Baselines}
\noindent \textbf{Multi-Agent End-to-End Methods.} 
We evaluate our proposed \textit{TurboTrain} paradigm by integrating it into multiple state-of-the-art (SOTA) multi-agent end-to-end models that employ spatiotemporal fusion for cooperative perception and prediction. To ensure consistency, we adopt the fusion framework from \cite{zhou2024v2xpnp} and implement various spatial fusion approaches, including \textit{FFNet} \cite{ffnet}, \textit{DiscoNet} \cite{disconet}, \textit{CoBEVFlow} \cite{wei2023asynchronyrobust}, and \textit{F-Cooper} \cite{fcooper}, all equipped with a baseline temporal module comprising alternating 2D and 3D convolutions. We also implement the \textit{V2XPnP} model \cite{zhou2024v2xpnp} and \textit{No Multi-Agent Fusion} baseline with an attention-based temporal module for comparison. Models marked with $^*$ indicate our re-implementations to ensure consistency in the LiDAR backbone (SECOND \cite{second}), temporal fusion network, and decoder heads.

\noindent \textbf{Pretrain Methods.} 
For multi-agent pretraining, we implement two strong SOTA self-supervised methods: \textit{CooPre} \cite{CooPre} and \textit{CORE} \cite{coreICCV}. Moreover, for single-agent pretraining, we employ \textit{T-MAE} \cite{wei2024tmae} as the SOTA temporal pretraining approach. We also include \textit{BEV-MAE} \cite{lin2024bevmae} as a reconstruction-based pretraining baseline and \textit{BEVContrast} \cite{bevcontrast2024} as a contrastive pretraining baseline.

\noindent \textbf{Balance MTL Methods.} 
We consider the following SOTA multi-task learning methods: 1) \textit{MoCo} \cite{fernando2023mitigating}: stochastic multi-objective gradient correction, 2) \textit{CAGrad} \cite{liu2021conflict}: conflict-averse gradient descent, which regularizes optimization based on the worst local improvement among tasks, 3) \textit{GradNorm} \cite{chen2018gradnorm}: normalization of task gradients to maintain a consistent gradient scale across multiple tasks.

\begin{table}[t]
  \caption{Comparison with SOTA pretraining methods on V2XPnP-Seq-VC dataset with V2XPnP model.}
  \label{tab:ablation_other_pretrain}
  \centering
  \footnotesize
  \renewcommand{\arraystretch}{1.15}
  \setlength{\tabcolsep}{3.5pt}
  {\begin{tabular}{l|c|ccc|c}
    \toprule[1.1pt]
    Methods & \textbf{AP@0.5} $\uparrow$ & ADE $\downarrow$ & FDE $\downarrow$ & MR $\downarrow$ & \textbf{EPA} $\uparrow$  \\
    \midrule 
    No Pretrain & 39.0 & 1.42 & 2.44 & 30.0 & 15.0 \\
    BEV-MAE \cite{lin2024bevmae} & 45.0 & 1.91 & 3.28 & 35.8 & 23.7 \\
    BEVContrast \cite{bevcontrast2024} & 50.3 & 1.79 & 3.16 & 39.8 & 27.1 \\
    T-MAE \cite{wei2024tmae} & 61.2 & 1.51 & 2.69 & 36.3 & 37.5 \\
    CORE \cite{coreICCV} & 63.0 & 1.50 & 2.74 & 36.4 & 38.6 \\
    CooPre \cite{CooPre} & 64.8 & 1.54 & 2.76 & 37.0 & 39.6 \\
    \rowcolor{gray!20} \textbf{TurboTrain (Pretrain)} & \textbf{70.3} & 1.54 & 2.80 & 37.2 & \textbf{43.4} \\
  \bottomrule[1.1pt]
\end{tabular}}
\vspace{-0.3cm}
\end{table}

\begin{table}[t]
  \caption{Comparison with SOTA balance MTL methods on V2XPnP-Seq-VC dataset with V2XPnP pretrained model. }
  \label{tab:ablation_other_Balance}
  \centering
  \footnotesize
  \renewcommand{\arraystretch}{1.1}
  \setlength{\tabcolsep}{4pt}
  {\begin{tabular}{l|c|ccc|c}
    \toprule[1.1pt]
    Methods & \textbf{AP@0.5} $\uparrow$ & ADE $\downarrow$ & FDE $\downarrow$ & MR $\downarrow$ & \textbf{EPA} $\uparrow$  \\
    \midrule 
    No Balance & 70.3 & 1.54 & 2.80 & 37.2 & 43.4 \\ 
    GradNorm \cite{chen2018gradnorm} & 60.9 & 1.73 & 3.18 & 40.6 & 34.2 \\
    MoCo \cite{fernando2023mitigating} & 66.0 & 1.79 & 3.18 & 40.3 & 38.2  \\
    CAGrad \cite{liu2021conflict} & \textbf{73.1} & 1.61 & 2.93 & 38.1 & 44.1 \\
    \rowcolor{gray!20} \textbf{TurboTrain (Balance)} & 72.2 & 1.49 & 2.75 & 35.0 & \textbf{45.5} \\
  \bottomrule[1.1pt]
\end{tabular}}
\vspace{-0.3cm}
\end{table}

\subsection{Main Results}
We investigate the effectiveness of \textit{TurboTrain} in achieving efficient and balanced multi-task learning for multi-agent perception and prediction. Notably, since prediction performance is dependent on detection quality, the varying difficulty of detecting different objects can significantly impact prediction metrics. Thus, \ul{\textit{EPA metric is the most important indicator for evaluating overall model performance}}. The main experiment results are shown in \cref{tab:main_results}. 

\noindent \textbf{\textit{TurboTrain} enhances multi-task learning for multi-agent perception and prediction.} 
TurboTrain demonstrates significant performance improvements over manual training strategies in cooperative perception and prediction tasks. Our method consistently achieves higher AP and EPA scores, where a higher EPA signifies superior detection and prediction capabilities. In the VC setup, TurboTrain-optimized V2XPnP achieves an AP of 72.2 and an EPA of 45.5, which are significantly improved compared to the manual training strategy. Furthermore, TurboTrain displays strong generalizability across different models and collaboration modes, indicating its broad applicability in end-to-end multi-agent systems. \cref{fig:comparison} presents the component analysis of \textit{Pretrain} stage and \textit{Balance} stage. \cref{fig:qualitative_results} provides a qualitative comparison of multi-task performance by different training strategies. More quantitative and qualitative results are provided in the Supplementary Material.

\begin{figure}[t]
    \centering
    \includegraphics[width=\linewidth]{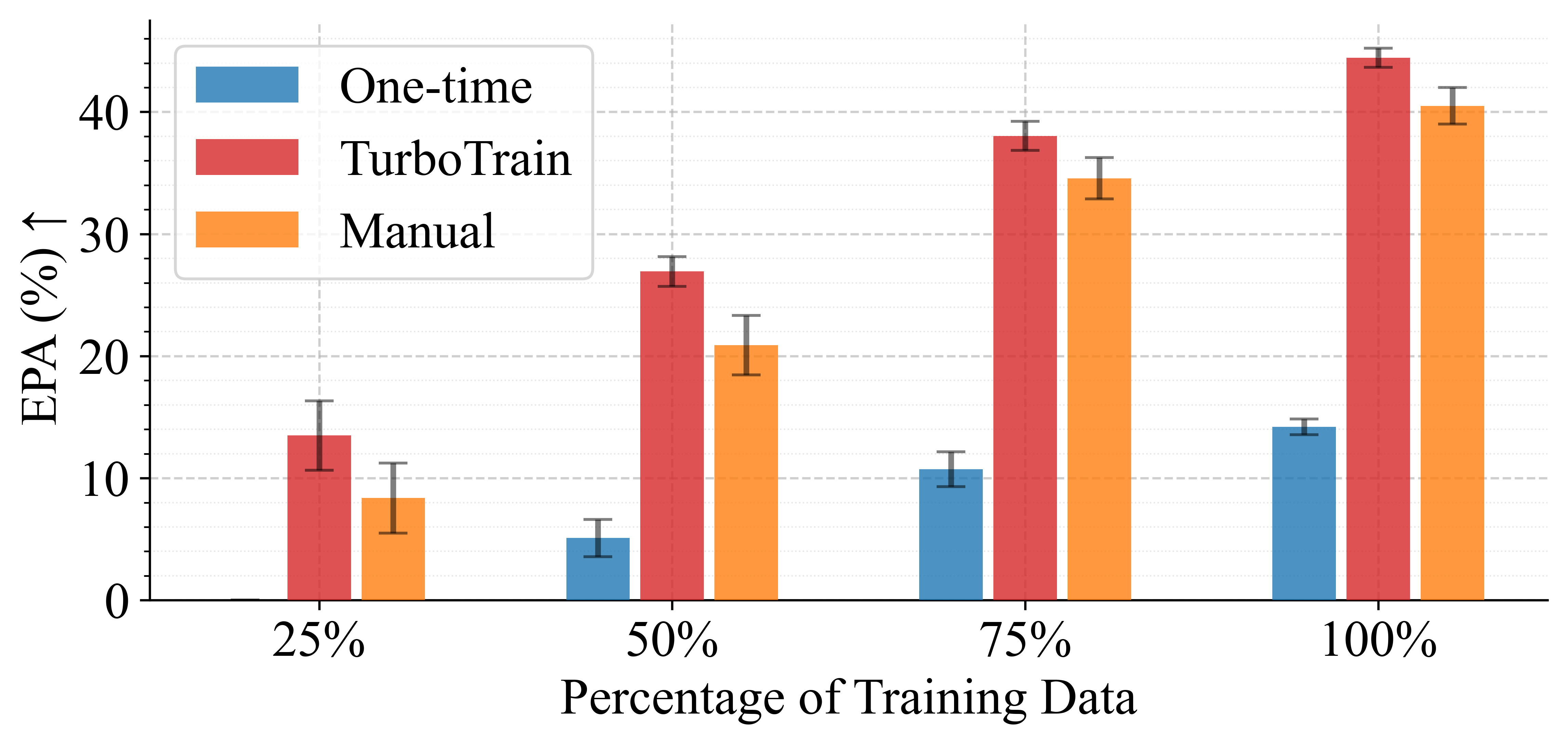}
    \caption{Data efficiency and training stability analysis under different training strategies. \textit{TurboTrain} consistently shows better performance with less variance among experiments. Note that one-time training fails when trained with 25\% of the full dataset. }
    \label{fig:data_efficiency}
    \vspace{-0.3cm}
\end{figure}

\begin{figure*}[tbh]
    \centering
    \includegraphics[width=\linewidth]{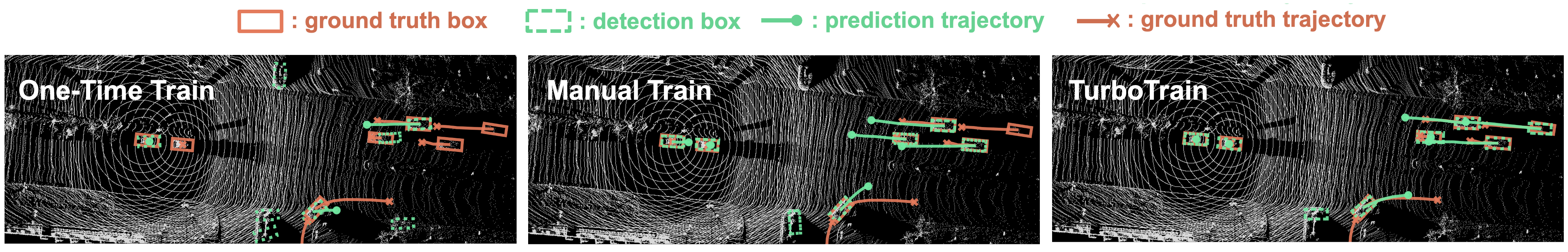}
    \caption{Qualitative comparison of different training strategies applied to the V2XPnP model for multi-agent perception and prediction tasks. The proposed \textit{TurboTrain} framework significantly enhances both detection and prediction quality, yielding more accurate results.}
    \label{fig:qualitative_results}
    \vspace{-0.2cm}
\end{figure*}

\noindent \textbf{\textit{TurboTrain} improves data efficiency and training stability.} 
To assess the benefits of our method in data-scarce scenarios, we employ V2XPnP \cite{zhou2024v2xpnp} as the backbone and randomly sample 25\%, 50\%, and 75\% of the training dataset. For our \textit{TurboTrain} method, the model is first pretrained on the full dataset and then fine-tuned within the \textit{Balance} stage on each sampled subset. Each experiment is conducted with three different random seeds, and we report the mean and standard deviation. As shown in \cref{fig:data_efficiency}, our method consistently outperforms the one-time training baseline across all settings. Notably, the performance improvement is more pronounced under data limitations. Furthermore, our approach exhibits a significantly lower standard deviation compared to the baseline, indicating enhanced training stability compared to other training methods.

\noindent \textbf{The learned 4D representation from the \textit{Pretrain} stage benefits the multi-agent end-to-end framework.} 
The learned 4D representation in \textit{Pretrain} significantly enhances the multi-agent end-to-end framework, as illustrated in \cref{fig:comparison}. Our \textit{Pretrain (No Balance Tuning)} approach effectively mitigates learning failures commonly associated with complex spatiotemporal features across multiple agents and frames. Even without the \textit{Balance} training stage, our approach achieves comparable performance to a manual training strategy, improving learning efficiency by reducing the need for extensive manual monitoring and training epochs. To further illustrate the importance of multi-agent, multi-frame pretraining, we compare our method against prior works, as shown in \cref{tab:ablation_other_pretrain}. For a fair comparison, we adopt a one-time training strategy without the \textit{Balance} stage. The results reveal that methods lacking temporal representation learning (\ie, CooPre, CORE, BEV-MAE) or multi-agent representation learning (\ie, T-MAE, BEV-MAE, BEVContrast) fail to match our method's superior performance, underscoring the necessity of incorporating both spatiotemporal and multi-agent representations.

\noindent \textbf{The \textit{Balance} stage facilitates multi-task learning.} 
The results in \cref{tab:ablation_other_Balance} indicate that our \textit{Balance} stage plays a critical role in adapting and optimizing individual tasks from task-agnostic features, as evidenced by the 4.8\% improvement over the no-balance setting. GradNorm primarily addresses gradient dominance across multiple tasks but results in a decline in EPA performance. In contrast, CAGrad leverages convex optimization to mitigate gradient conflicts and improve EPA. MoCo considers both factors but achieves only moderate performance, suggesting that the primary challenge in multi-agent prediction and detection stems from conflict rather than dominance. The superior performance of our conflict-suppressing in \textit{TurboTrain} further highlights the impact of task conflicts and achieves a better overall balance in multi-task learning.

\subsection{Ablation Studies}
In this section, we present ablation studies that evaluate the impact of our key technical components. Additional ablation results are provided in the Supplementary Material.

\begin{table}[t]
\centering
\captionsetup{width=0.45\linewidth}
\begin{minipage}[t]{0.5\linewidth} 
\centering
\footnotesize
\caption{Influence of Multi-agent Fusion Integration in \textit{Pretrain} stage on V2XPnP-Seq-VC dataset.}
\setlength{\tabcolsep}{4pt}
\begin{tabular}{c|c|c}
    \toprule
    Method & \textbf{AP@0.5} $\uparrow$ & \textbf{EPA} $\uparrow$  \\
    \midrule 
    No Pretrain & 39.0 & 15.0 \\
    wo Mul. Fus. & 65.1 & 40.8 \\
    w/ Mul. Fus. & 70.3 & 43.4 \\
    \bottomrule
\end{tabular}
\label{tab:ablation_module_init}
\end{minipage}%
\hfill 
\begin{minipage}[t]{0.5\linewidth}
\centering
\footnotesize
\setlength{\tabcolsep}{4pt}
\caption{Analysis on Reconstruction Loss with V2XPnP model in \textit{Pretrain} stage on V2XPnP-Seq-VC dataset.}
\begin{tabular}{c|c|c}
    \toprule
    Loss & \textbf{AP@0.5} $\uparrow$ & \textbf{EPA} $\uparrow$ \\
    \midrule 
    $\mathcal{L}_{rec}$ & 65.8 & 40.2 \\ 
    $\mathcal{L}_{occ}$ & 66.3 & 40.8 \\
    Both & 70.3 & 43.4 \\
    \bottomrule
\end{tabular}
\label{tab:ablation_recon_loss}
\end{minipage}
\vspace{-0.4cm}
\end{table}

\noindent \textbf{Ablation on Multi-agent Module Integration in \textit{Pretrain} Stage.} 
We evaluate the impact of integrating the multi-agent fusion module during pretraining, as shown in \cref{tab:ablation_module_init}. The results indicate that excluding the multi-agent fusion module during pretraining leads to learning failure on V2XPnP, underscoring the necessity of the module.

\noindent \textbf{Ablation on Reconstruction Loss in \textit{Pretrain} Stage.} 
\cref{tab:ablation_recon_loss} presents the performance comparison under different reconstruction loss configurations. We find that leveraging both losses yields the highest performance, highlighting the importance of incorporating both reconstruction objectives to enhance feature learning.

\noindent \textbf{Influence of Parameter Settings in \textit{Balance} Stage.} 
The results in \cref{tab:ablation_Balance_Parameter} reveal that applying balance at every gradient step, especially from the very beginning, leads to training failure. This is because, during the initial training phase, task-specific heads require a degree of random exploration. Premature balancing constraints this exploration, limiting their capacity to learn and potentially discard valuable pre-trained features. Moreover, implementing gradient balancing requires additional backward computations for each task-specific head, which significantly increases GPU memory consumption (1.5× times) and prolongs training time. Our hybrid free and conflict-suppressing strategy with $(n,m)=(2000,1000)$ achieves the best performance while maintaining a low training cost. A smaller interval misleads the training process, while a larger interval causes the balance mechanism to degrade.

\begin{table}[t]
  \caption{Ablation studies of the iteration parameter $n, m$ (free training step number, balance step number) on V2XPnP-Seq-VC dataset with V2XPnP pretrained model.}
  \label{tab:ablation_Balance_Parameter}
  \centering
  \footnotesize
  \renewcommand{\arraystretch}{1.15}
  {\begin{tabular}{l|c|ccc|c}
    \toprule[1.1pt]
    $n, m$ & \textbf{AP@0.5} $\uparrow$ & ADE $\downarrow$ & FDE $\downarrow$ & MR $\downarrow$ & \textbf{EPA} $\uparrow$ \\
    \midrule 
    Bal Per-step & 0.0 & - & - & - & - \\ 
    1000, 500 & 65.8 & 1.67 & 3.06 & 41.4 & 38.1  \\
    \rowcolor{gray!20} \textbf{2000, 1000} & \textbf{72.2} & 1.49 & 2.75 & 35.0 & \textbf{45.5} \\
    3000, 1500 & 70.2 & 1.53 & 2.80 & 37.8 & 42.8 \\
  \bottomrule[1.1pt]
\end{tabular}}
\vspace{-0.2cm}
\end{table}

\section{Conclusion}
In this work, we introduce \textbf{TurboTrain}, a novel learning framework designed to address the challenges of end-to-end multi-task learning in multi-agent autonomous driving systems. TurboTrain eliminates the need for complex multi-stage training pipelines while significantly improving cooperative detection and prediction performance. Our task-agnostic pretraining strategy effectively captures spatiotemporal dependencies, enhancing feature learning across different tasks. Additionally, our gradient-alignment balancer mitigates multi-task conflicts, ensuring stable and efficient training without excessive computational overhead. Through extensive evaluations on a large-scale real-world V2X dataset, we demonstrate that TurboTrain further improves the state-of-the-art cooperative methods in end-to-end perception and prediction tasks.

\section*{Acknowledgements}

This work was supported by the Federal Highway Administration Center of Excellence on New Mobility and Automated Vehicles, and by the National Science Foundation under Award No. 2346267, POSE: Phase II - DriveX: An Open Source Ecosystem for Automated Driving and Intelligent Transportation Research.

\nocite{lang_pointpillars_2019, cosineanneal, kingma2014adam, cooperfuse}
{
    \small
    \bibliographystyle{ieeenat_fullname}
    \bibliography{main}
}

\addtocontents{toc}{\protect\setcounter{tocdepth}{2}} 
\clearpage
\maketitlesupplementary

\renewcommand{\thesection}{\Alph{section}}
\setcounter{section}{0}

\setcounter{table}{0}
\renewcommand{\thetable}{S\arabic{table}}

\setcounter{figure}{0}
\renewcommand{\thefigure}{S\arabic{figure}}

\setcounter{equation}{0}
\renewcommand{\theequation}{S\arabic{equation}}

\section{Details of Problem Formulation}
\subsection{Explanation of Different Training Strategies}

In Sec. 3 of the main paper, we show the comparison of one-time training, manual training, and \textit{TurboTrain} in Fig.2. The numerical experiment results of three training strategies are provided in \cref{tab:supple_prelim}.

\noindent \textbf{One-time training strategy}. This strategy directly trains all the modules in the end-to-end framework. Row 1 (One-time (s-f)) indicates we only train \textit{multi-agent single-frame perception}. Row 2 (One-time (s-a)) indicates we only train \textit{single-agent perception and prediction}. Row 3 (One-time (all)) indicates we only train \textit{multi-agent perception and prediction}. 

\noindent \textbf{Manual training}. Rows four to seven indicate the performance breakdown of the manual training strategy. For the manual training strategy, we divide the training process into four stages. \textit{Stage 1: Single-Agent Detection} (Row 4). We train the detection backbone on single frames to build a robust foundation for object recognition. \textit{Stage 2: Single-agent Temporal Prediction} (Row 5). We freeze the detection backbone and train a dedicated temporal network and prediction head to capture complex temporal dynamics for the prediction task. \textit{Stage 3: Single-agent Perception and Prediction Joint Fine-Tuning} (Row 6). We unfreeze the entire network and fine-tune all components end-to-end, harmonizing the perception and prediction tasks. \textit{Stage 4: Multi-Agent Fusion} (Row 7). We incorporate a multi-agent fusion module with dynamic loss weighting to jointly optimize spatiotemporal representations across agents, ensuring balanced performance. 

\noindent \textbf{TurboTrain}. Row 8 indicates that we only use the Pretrain stage of TurboTrain and finetune the model without using the Balance stage. Row 9 indicates we use the whole TurboTrain pipeline. From the results, we observe that 1) one-time training faces challenges with multi-agent  spatiotemporal feature learning; 2) manual training strategy avoids such learning failure but requires more monitoring stages; 3) TurboTrain achieves superior performance while avoiding such learning inefficiency.

\begin{table}[t]
  \caption{Comparison of One-time training, Manual training, and \textit{TurboTrain} on the V2XPnP-Seq-VC dataset with the V2XPnP model. {\textcolor{BurntOrange}{-x}} represents the performance decline compared to the manual training strategy, and {\textcolor{darkgreen}{+x}} represents the improvement. In one-time training strategy, "s-f" indicates the single frame model, "s-a" means the single agent model, and "all" represents the multi-frame multi-agent model. Moreover, "No Bal" represents a pretrain-only model without the balance stage.}
  \label{tab:supple_prelim}
  \centering
  \scriptsize
  \renewcommand{\arraystretch}{1.15}
  \setlength{\tabcolsep}{1.8pt}
  {\begin{tabular}{l|c|ccc|c}
    \toprule[1.1pt]
    Strategy & \textbf{AP@0.5}(\%) $\uparrow$ & ADE(m) $\downarrow$ & FDE(m) $\downarrow$ & MR(\%) $\downarrow$ & \textbf{EPA}(\%) $\uparrow$  \\
    \midrule 
    One-time (s-f)  & 68.3\tiny{\textcolor{BurntOrange}{-2.0}} & - & - & - & - \\ 
    One-time (s-a)  & 55.1\tiny{\textcolor{BurntOrange}{-15.2}} & 1.60 & 2.99 & 37.5 & 29.9\tiny{\textcolor{BurntOrange}{-12.9}} \\ 
    One-time (all) & 39.0\tiny{\textcolor{BurntOrange}{-31.3}} & 1.42 & 2.44 & 30.0 & 15.0\tiny{\textcolor{BurntOrange}{-27.8}} \\ 
    \midrule 
    Single-agent Det & 45.2 & - & - & - & -  \\
    Temporal Pred & 48.2 & 2.03 & 3.55 & 38.3 & 20.2 \\
    Joint Tuning & 57.4 & 1.58 & 2.89 & 38.3 & 32.5 \\
    Multi-agent Fus  & 70.3 & 1.53 & 2.80 & 37.8 & 42.8  \\
    \midrule 
    TurboTrain (No Bal) & 70.3 & 1.54 & 2.80 & 37.2 & 43.4 \\
    \rowcolor{gray!20} \textbf{TurboTrain} & \textbf{72.2}\tiny{\textcolor{darkgreen}{+1.9}} & \textbf{1.49} & \textbf{2.75} & \textbf{35.0} & \textbf{45.5}\tiny{\textcolor{darkgreen}{+2.7}} \\
  \bottomrule[1.1pt]
\end{tabular}}
\end{table}

\subsection{Motivating Research Questions}
\noindent \textbf{Q: Why is it challenging for efficient and balanced multi-task learning for multi-agent perception and prediction and why is such a problem important?}

\noindent \textbf{A:} Training a system that processes data from multiple agents across several frames presents significant challenges. Conventional one-time training, where all tasks are learned jointly from scratch, fails to capture the intricate features that arise from merging temporal and multi-agent data. This is due to the inherent complexity of integrating spatial and temporal information, which often leads to unstable training and suboptimal performance. For instance, tasks like detection may dominate, thereby overshadowing others such as trajectory prediction. As demonstrated in \cref{tab:supple_prelim}, configurations involving either single-agent multi-task or multi-agent single-task scenarios do not encounter these issues, whereas the multi-agent multi-task setup places much higher demands on the feature learning process. Moreover, manual multi-stage training strategies heavily depend on the careful selection of checkpoints at each stage, and system errors tend to accumulate as the number of stages increases. This approach also relies on annotated labels, making it less effective in scenarios where data is scarce. These challenges highlight the need to thoroughly investigate and address this learning problem. The following questions further clarify our motivation for developing the solutions presented in our work.

\noindent \textbf{Q: Why use mask reconstruction pretraining?}

\noindent \textbf{A:} We adopt mask reconstruction pretraining to enable our model to learn robust features without relying on annotated labels. This approach generates a comprehensive 4D representation, which serves as a strong initialization for critical components in our end-to-end framework—namely, the 3D Detection Encoder, Temporal Fusion Module, and Multi-agent Fusion Module. As demonstrated in our main paper, each of these modules plays an essential role in system performance. By masking portions of the temporal input data and requiring the network to reconstruct them for each agent, we compel the model to learn the scene's underlying structure as well as sensor distributions of heterogeneous agents. Our dual reconstruction strategy, which operates at both the point level and voxel level, allows the pretrained module to capture detailed 3D structures for the perception task and to understand scene occupancy layouts for the prediction task. This methodology proves particularly valuable when data is scarce or incomplete, as it enables the model to extract fine details at multiple scales, improving its ability to handle occlusions and missing data—common challenges in real-world driving scenarios. Furthermore, our experiments confirm the effectiveness of this approach in data-scarce environments.

\noindent \textbf{Q: Why use balancing in multi-task learning?}

\noindent \textbf{A:} Balancing is critical in our multi-task setup. Different tasks (like detection and prediction) can produce conflicting gradient signals during training, meaning one task improves while another is worse. Our approach uses a conflict-suppressing gradient alignment mechanism to remove the conflict components between multiple tasks. This helps stabilize training and improves overall performance by mitigating the negative cross-task interference. Moreover, our hybrid training strategy alternates between free and balanced gradient steps, making the training process more efficient without a heavy computational cost.

\section{Implementation Details}
\label{app:implementation}
In this section, we provide detailed configurations for our \textit{TurboTrain} paradigm. Moreover, we go further to the baseline models used in our experiments for cooperative perception and prediction tasks.

\subsection{Baseline Model Details}

\noindent \textbf{LiDAR Perception Backbone.} 
We adopt the anchor-based SECOND model \cite{second} as the LiDAR feature extraction backbone for all the baseline models. The voxel resolution in the experiment is set to 0.1 m in both the $x$ and $y$ directions and 0.2 m in the $z$ direction, with a maximum of 5 points per voxel and up to 32,000 voxels. Additionally, we configure 2 anchors per BEV grid cell.

\noindent \textbf{Multi-Agent End-to-End Methods.} 
As research into end-to-end multi-agent perception and prediction tasks is still in the early stage, most existing studies primarily focus on single-frame spatial fusion or incorporating short-term temporal information. Therefore, we adopt the V2XPnP end-to-end framework and benchmark \cite{zhou2024v2xpnp} to reimplement and evaluate state-of-the-art (SOAT) V2X fusion methods. Specifically, we follow the V2XPnP framework and leverage the baseline temporal fusion model of V2XPnP to build the end-to-end spatiotemporal fusion framework for multiple tasks and incorporate SOAT multi-agent fusion methods as baselines, including \textit{FFNet} \cite{ffnet}, \textit{DiscoNet} \cite{disconet}, \textit{CoBEVFlow}, \textit{F-Cooper} \cite{fcooper}, and V2XPnP \cite{zhou2024v2xpnp}. From the main experiment results in Sec. 5, we demonstrate that our \textit{TurboTrain} method is generalizable across various fusion methods.

\noindent \textbf{Map Feature Extraction.} 
HD maps for prediction are modeled as sets of polylines, where each map polyline comprises 10 sequential points. For BEV projection, each grid cell retains the five nearest polylines to reduce redundancy. Each waypoint is defined by seven attributes: $(x, y, d_x, d_y, type, x_{pre}, y_{pre})$, which captures its spatial coordinates, orientation, lane type, and preceding position. These attributes are embedded via MLP layers and the map interaction is captured by one Transformer layer.

\begin{figure*}[tbh]
    \centering
    \includegraphics[width=0.98\linewidth]{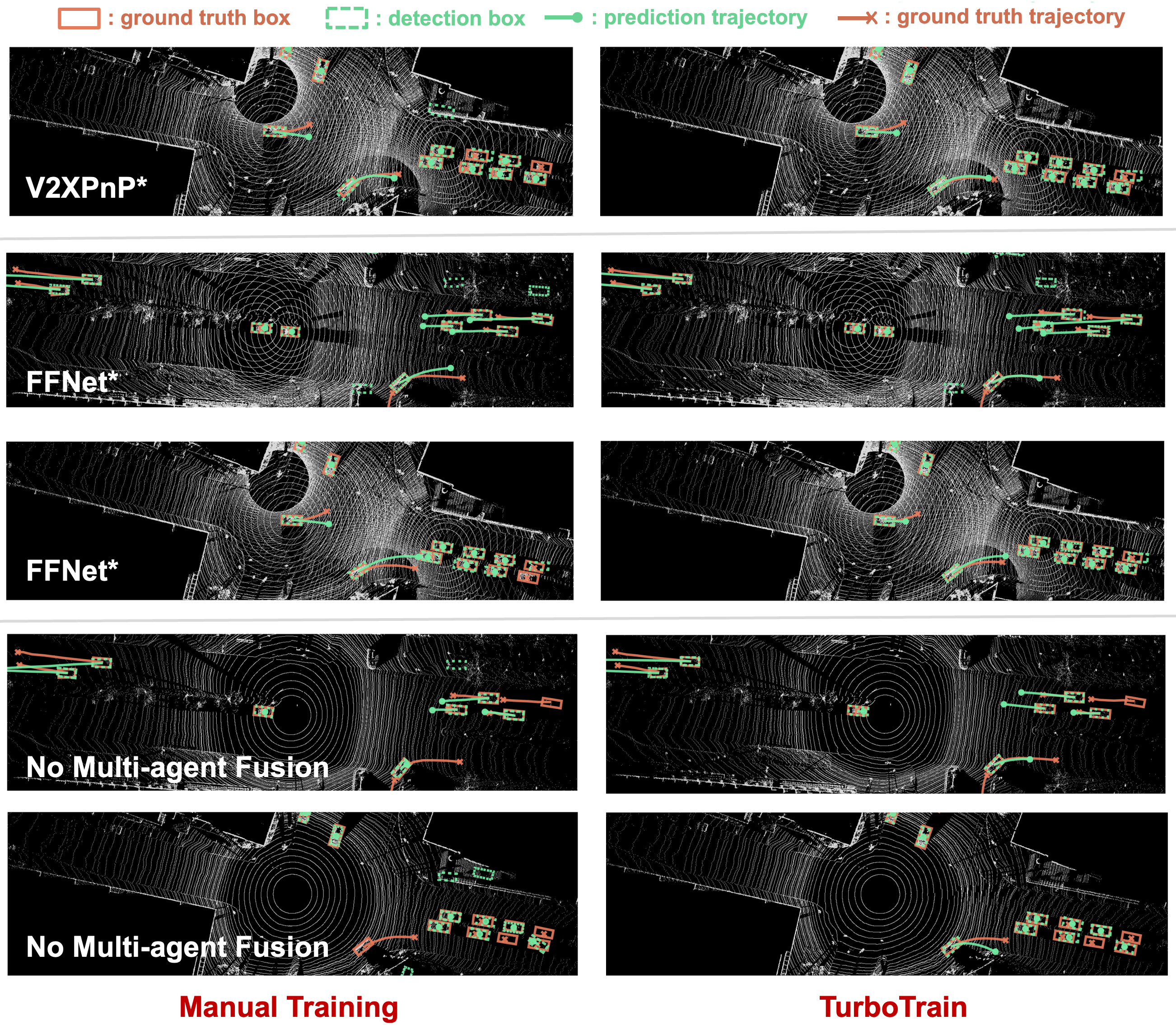}
    \caption{Qualitative comparison of different training strategies applied to the different fusion methods for multi-agent perception and prediction tasks. The proposed \textit{TurboTrain} framework significantly enhances both detection and prediction quality over different fusion methods. Models marked with $^*$ indicate our reimplementations to ensure consistency with prior works.}
    \label{fig:supple_qualitative_results}
\end{figure*}

\subsection{Additional Pretraining Details}

\noindent \textbf{Formulation of occupancy reconstruction loss.} Given the predicted occupancy value and the ground truth occupancy value, we define the occupancy prediction loss as:
\begin{equation}
\mathcal{L}_{occ} = - \alpha \left(1 - Pr^{i}\right)^{\gamma} \log(Pr^{i}),
\end{equation}
where $Pr^{i}$ represents the predicted probability of voxel $i$. The weighting factor $\alpha$ is set to 2, and the weighting factor $\gamma$ is set to 0.25.

\subsection{Detailed Experimental Setting}
\noindent \textbf{Training details.} The pretraining network contains a 3D Backbone, Temporal Fusion module, Multi-agent Fusion Module, and two separate lightweight decoder heads for reconstruction tasks. During the finetuning stage, only the pretrained 3D Backbone, Temporal Fusion module and Multi-agent Fusion Module will be taken. During pretraining, we employ AdamW \cite{kingma2014adam} optimizer with a weight decay of $1\times10^{-2}$ to optimize our models. We train the model with a batch size of 4 for 15 epochs using a learning rate of 0.002, and we decay the learning rate with a cosine annealing \cite{cosineanneal}. We use a masking ratio of 0.7 in our main experiments and a fixed predicted point cloud number of 20 for point cloud reconstruction. During the fine-tuning stage, the optimization process is identical to the train-from-scratch baselines. 

\noindent \textbf{Testing details.} During testing, a fixed agent is designated as the ego agent in each cooperative scenario, while during training, the ego agent is randomly assigned. Following the real-world evaluation setting \cite{cooperfuse}, the communication range is set to $50$ meters with surrounding agents evaluated within the range of $x\in[-102.4, 102.4]$ m and $y\in[-40,40]$ m, and messages exceeding $50$ meters are discarded. The historical observation length is set to $2$ seconds ($2$ Hz), while the prediction horizon extends to $3$ seconds ($2$ Hz).

\section{Additional Results}

\subsection{Ablation Studies}

\noindent \textbf{Ablation on Reconstruction Objective Design.} 
We evaluate the impact of different reconstruction objectives by varying the number of historical frames used, as shown in \cref{tab:ablation_temporal}. Specifically, we compare reconstructing from all available historical frames versus using only half. Given that the historical observation window is set to 2 seconds, our results indicate that leveraging a larger number of past frames substantially enhances the model’s performance, highlighting the importance of temporal context in representation learning.

\noindent \textbf{Ablation on Masking ratio.} 
We investigate the effect of varying the masking ratio and observe that a ratio of 0.7 yields the best performance, as shown in \cref{tab:ablation_mask_ratio}. This suggests that an optimal level of information removal is crucial for effective learning, balancing the trade-off between preserving sufficient context and encouraging robust feature extraction.

\begin{table}[tbh]
  \caption{Ablation on Input Temporal Frames on V2XPnP-Seq-VC dataset with V2XPnP model. $T$ denotes the total number of historical frames. }
  \label{tab:ablation_temporal}
  \centering
  \begin{tabular}{c|c|c}
    \toprule
    Input Frames & \textbf{AP@0.5} $\uparrow$ & \textbf{EPA} $\uparrow$ \\
    \midrule 
    $T/2$ & 64.2 & 38.8 \\ 
    $T$ &  \textbf{70.3} & \textbf{43.4} \\
    \bottomrule
\end{tabular}
\end{table}

\begin{table}[tbh]
    \centering
    \caption{Ablation on Reconstruction Mask Ratio on V2XPnP-Seq-VC dataset with V2XPnP model.}
    \begin{tabular}{c|cc}
    \toprule[1.1pt]
    Mask Ratio & \textbf{AP@0.5} $\uparrow$ & \textbf{EPA} $\uparrow$  \\
    \midrule 
    0.7 & \textbf{70.3} & \textbf{43.4} \\ 
    0.8 &  67.1 & 42.9 \\
    0.9 & 66.4 & 42.1 \\
    \bottomrule[1.1pt]
    \end{tabular}
    \label{tab:ablation_mask_ratio}
\end{table}

\subsection{Qualitative Results}
\cref{fig:supple_qualitative_results} shows additional qualitative comparisons across vehicle-centric and vehicle-to-vehicle scenario with various models: (1) No Multi-Agent Fusion, (2) FFNet \cite{ffnet}, and (3) V2XPnP \cite{zhou2024v2xpnp}. Our TurboTrain framework consistently improves both detection and prediction performance across these configurations.

\end{document}